\documentstyle[aclap]{article}

\title{\vspace{-0.5in}Memoization of Coroutined Constraints}

\author{
\hbox{\begin{tabular}{c}
  Mark Johnson\\
  \normalsize Cognitive and Linguistic Sciences, Box 1978\\
  \normalsize Brown University\\
  \normalsize Providence, RI 02912, USA\\
  {\normalsize \sf Mark\_Johnson@Brown.edu}
\end{tabular}}
\hbox{\begin{tabular}{c}
  Jochen D\"orre\thanks{This research was largely conducted at
the Institut f\"ur maschinelle Sprachverarbeitung in Stuttgart.
We would like to thank Andreas Eisele, Pascal van Hentenryck, Martin Kay,
Fernando Pereira, Edward Stabler and our colleagues at the
Institut f\"ur maschinelle Sprachverarbeitung for helpful comments
and suggestions. All remaining errors are our own.  The
Prolog code presented in this paper is available via anonymous ftp
 from {\sf lx.cog.brown.edu} as {\sf /pub/lemma.tar.Z}} \\
  \normalsize Institut f\"ur maschinelle Sprachverarbeitung\\
  \normalsize Universit\"at Stuttgart\\
  \normalsize D--70174 Stuttgart, Germany\\
  {\normalsize \sf Jochen.Doerre@ims.uni--stuttgart.de}
\end{tabular}}}

\font\ninett=cmtt10 at 9pt

\def\beginprolog{\par\vspace{-.5ex}\begingroup\leftskip=0.05in\ninett
                 \setlength{\parindent}{0in}
                 \setlength{\parskip}{0in}
                 \def\par{\leavevmode\endgraf}  
                 \catcode`\^=12 
                 \catcode`\&=12 %
                 \catcode`\_=12 %
                 \catcode`\|=12 %
                 \obeylines%
                 \catcode`\ =\active%
                 \catcode`\%=12\catcode`\`=\active}
{\catcode`\ =\active\global\let =\enskip}
{\catcode`\`=\active\gdef`{\relax\lq}}

\def\endprolog{\endgroup\par\vspace{2.2ex}}

\def\xbar{$X'$ }

\def\ltatom#1{$\rm #1$}
\def\ltulatom#1{$\underline{\rm #1}$}

\def\ltbclause#1#2{#1 $\leftarrow$ #2.}
\def\ltprefix#1{\hbox to 1.75cm{\small #1} $\;$}
\def\lttable#1#2#3#4{\hbox{\ltprefix{#1.#2[#3]\hfil {\footnotesize T}} #4}\par}
\def\ltprogram#1#2#3#4{\hbox{\ltprefix{#1.#2[#3]\hfil {\footnotesize P}}
#4}\par}
\def\ltsolution#1#2#3#4{\hbox{\ltprefix{#1.#2[#3]\hfil {\footnotesize S}}
#4}\par}

\def\sqr#1#2{{\vcenter{\vbox{\hrule height.#2pt
        \hbox{\vrule width.#2pt height#1pt \kern#1pt
           \vrule width.#2pt}
        \hrule height.#2pt}}}}
\def\Box{\mathchoice\sqr44\sqr44\sqr{2.1}3\sqr{1.5}3}

\newcommand{\la}{\leftarrow}

\newtheorem{Theorem}{Theorem}

\newtheorem{Definition}[Theorem]{Definition}

\newcommand{\clause}[2]{{#1 \leftarrow #2}}

\newcommand{\cvar}[1]{#1}

\newcommand{\program}{\mbox{\sf program}}
\newcommand{\tabelle}{\mbox{\sf table}}
\newcommand{\solution}{\mbox{\sf solution}}

\newcommand{\ul}{{\tt \_}}

\newcommand{\A}{{\cal A}}

\newcommand{\bnode}[2]{{{\displaystyle\strut #2}\over{\displaystyle\strut #1}}}
\newcommand{\tnode}[2]{{{{{\displaystyle\strut }\atop}\atop{\displaystyle\strut
                          #2}}\over{\displaystyle\strut #1}}}
\newcommand{\cat}[1]{\mbox{\footnotesize #1}}

\newcommand{\gp}[2]{\hbox{\vtop{\hbox{#1} \hbox{#2}}}}

\newcount\exno
\newcommand{\example}[1]{{\raggedright %
                          \begin{list}{\advance\exno by1 (\the\exno)}{} %
                           \item #1 %
                          \end{list}}}

\begin{document}

\maketitle

\begin{abstract}
Some linguistic constraints cannot be effectively resolved during
parsing at the location in which they are most naturally introduced.
This paper shows how constraints can be propagated in a memoizing parser
(such as a chart parser) in much the same way that variable bindings
are, providing a general treatment of constraint coroutining in
memoization.  Prolog code for a simple application of our technique to
Bouma and van Noord's (1994) categorial grammar analysis of Dutch is
provided.
\end{abstract}

\section{Introduction}

\noindent
As the examples discussed below show, some linguistic
constraints cannot be effectively resolved during parsing
at the location in which they are most naturally introduced.
In a backtracking parser, a natural way of dealing with such
constraints is to coroutine them with the other parsing processes,
reducing them only when the parse tree is sufficiently instantiated
so that they can be deterministically resolved.
Such parsers are particularly easy to implement in extended
versions of Prolog (such as PrologII, SICStus Prolog and Eclipse)
which have such coroutining facilities built-in.
Like all backtracking parsers, they
can exhibit non-termination and exponential parse times
in situations where memoizing parsers (such as chart parsers)
can terminate in polynomial time.
Unfortunately, the coroutining approach, which requires that
constraints share variables in order to communicate, seems to
be incompatible with standard memoization techniques,
which require systematic variable-renaming (i.e., copying)
in order to avoid spurious variable binding.

For generality, conciseness and precision,
we formalize our approach to memoization and
constraints within H\"ohfeld and Smolka's~(1988) general
\nocite{DR88}
theory of Constraint Logic Programming (CLP), but we discuss how our
method can be applied to more standard chart parsing as well.  This
paper extends our previous work reported in D\"orre~(1993) \nocite{DorreDyana}
and Johnson~(1993) \nocite{MJ93} by generalizing those methods to arbitrary
constraint systems (including feature-structure constraints), even
though for reasons of space such systems are not discussed here.

\section{Lexical rules in Categorial Grammar}

\noindent
This section reviews Bouma and van Noord's~(1994) (BN henceforth)
constraint-based
\nocite{BoumaNoord94a}
categorial grammar analysis of modification in Dutch, which we use
as our primary example in this paper.
However, the memoizing CLP interpreter presented below has also
been applied to GB and HPSG parsing, both of which benefit from
constraint coroutining in parsing.

BN can explain a number of puzzling scope phenomena by
proposing that
heads (specifically, verbs) subcategorize for adjuncts as well
as arguments (rather than allowing adjuncts to subcategorize
for the arguments they modify, as is standard in Categorial Grammar).
For example, the first reading of the Dutch sentence

\begin{figure*}
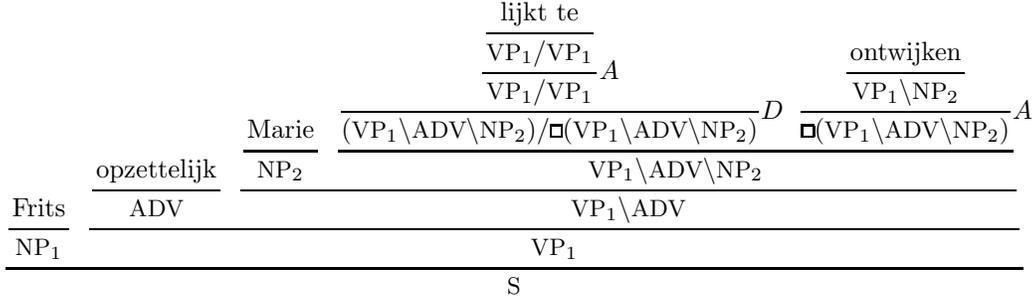

%
\[
\bnode{ \cat{S} }
{\tnode{\cat{NP}_1}{\rm Frits} \;\;
 \bnode{ \cat{VP}_1 }
 {\tnode{\cat{ADV}}{\rm opzettelijk} \;\;
  \bnode{ \cat{VP}_1 \backslash \cat{ADV} }
  {\tnode{\cat{NP}_2}{\rm Marie} \;\;
   \bnode{ \cat{VP}_1 \backslash \cat{ADV} \backslash \cat{NP}_2 }
   {\bnode{ (\cat{VP}_1 \backslash \cat{ADV} \backslash \cat{NP}_2) /
            \Box (\cat{VP}_1 \backslash \cat{ADV} \backslash \cat{NP}_2) }
          {\bnode{ \cat{VP}_1 / \cat{VP}_1 }
                 {\tnode{ \cat{VP}_1 / \cat{VP}_1 }
                          {\mbox{lijkt te}}}
                 A} D
    \;\;
    {\bnode{\Box(\cat{VP}_1 \backslash \cat{ADV} \backslash \cat{NP}_2)}
           {\tnode{ \cat{VP}_1 \backslash \cat{NP}_2 }
                  {\rm ontwijken}}}
    \rlap{$A$}}}}}
\]
\caption{The BN analysis of (1).  In this derivation
`$\protect\cat{VP}_1$' abbreviates `$\protect\cat{S} \backslash
\protect\cat{NP}_1$',
`$A$' is a lexical rule which adds adjuncts
to verbs, `$D$' is a lexical `division' rule which enables a control or raising
verb to
combine with arguments of higher arity, and `$\Box$' is a unary modal
operator which diacritically marks infinitival verbs.
\label{fig:cgder}}
\end{figure*}

\example{
Frits \gp{opzettelijk}{deliberately} Marie \gp{lijkt te}{seems}
\gp{ontwijken}{avoid} \break
\strut `Fritz deliberately seems to avoid Marie' \break
`Fritz seems to deliberately avoid Marie'}
\noindent
is obtained by the analysis depicted in Figure~\ref{fig:cgder}.
The other reading of this sentence is produced by a derivation
in which the adjunct addition rule `$A$' adds an adjunct to
{\em lijkt te}, and applies vacuously to {\em ontwijken}.

It is easy to formalize this kind of grammar in pure Prolog.
In order to simplify the presentation of the proof procedure interpreter below,
we write clauses as `$H$ \verb+::-+ $B$'
where $H$ is an atom (the head) and $B$ is a list of atoms (the negative
literals).

The atom $\verb+x(+{\it Cat, Left, Right}\verb+)+$ is true iff the substring
between the two string positions {\it Left} and {\it Right} can be
analyzed as belonging to category {\it Cat}.
(As is standard, we use
suffixes of the input string for string positions).

The modal operator `$\Box$' is used to diacritically mark untensed verbs
(e.g., {\it ontwijken}), and prevent them from combining with their
arguments.  Thus untensed verbs must combine with other verbs which
subcategorize for them (e.g., {\it lijkt te}), forcing all verbs to
appear in a `verb cluster' at the end of a clause.

For simplicity we have not provided a semantics here, but it is easy to
add a `semantic interpretation' as a fourth argument in the usual manner.
The forward and backward application rules are specified as clauses of
\verb+x+/3.  Note that the application rules are left-recursive,
so a top-down parser will in general fail to terminate with such
a grammar.
\beginprolog
:- op(990, xfx, ::- ).  
:- op(400, yfx, \verb+\+ ).    
:- op(300, fy, \# ).     

x(X, Left, Right) ::- [ 
    x(X/Y, Left, Mid),
    x(Y, Mid, Right) ].
x(X, Left, Right) ::- [ 
    x(Y, Left, Mid),
    x(X\verb+\+Y, Mid, Right) ].
x(X, [Word|Words], Words) ::- [
    lex(Word, X) ].
\endprolog
\noindent
Lexical entries are formalized using a two place relation
$\verb+lex(+{\it Word, Cat}\verb+)+$, which is true if {\it Cat} is a category
that the lexicon assigns to {\it Word}.
\beginprolog
lex('Frits', np) ::- [].
lex('Marie', np) ::- [].
lex(opzettelijk, adv) ::- [].
lex(ontwijken, \#X ) ::- [
    add_adjuncts(s\verb+\+np\verb+\+np, X ) ].
lex(lijkt_te, X / \#Y ) ::- [
    add_adjuncts((s\verb+\+np)/(s\verb+\+np), X0),
    division(X0, X/Y ) ].
\endprolog
\noindent
The \verb+add_adjuncts+/2 and \verb+division+/2 predicates formalize
the lexical rules `$A$' (which adds adjuncts to verbs) and `$D$' (the
division rule).
\beginprolog
add_adjuncts(s, s) ::- [].
add_adjuncts(X, Y\verb+\+adv) ::- [
    add_adjuncts(X, Y) ].
add_adjuncts(X\verb+\+A, Y\verb+\+A) ::- [
    add_adjuncts(X, Y) ].
add_adjuncts(X/A, Y/A) ::- [
    add_adjuncts(X, Y) ].

division(X, X) ::- [].
division(X0/Y0, (X\verb+\+Z)/(Y\verb+\+Z)) ::- [
    division(X0/Y0, X/Y) ].
\endprolog
\noindent
Note that the definitions of \verb+add_adjuncts+/2 and \verb+division+/2 are
recursive, and have an infinite number of solutions when only their first
arguments are instantiated.
This is necessary because the number of adjuncts that can be associated with
any given verb is unbounded.
Thus it is infeasible to enumerate all of the categories
that could be associated with a verb when it is retrieved from the lexicon,
so following
BN, we treat the predicates
\verb+add_adjuncts+/2 and \verb+division+/2 as coroutined constraints which
are only resolved when their second arguments become sufficiently instantiated.

As noted above, this kind of constraint coroutining is built-in to a number
of Prolog implementations.  Unfortunately, the left recursion inherent in
the combinatory rules mentioned earlier dooms any standard backtracking
top-down
parser to non-termination, no matter how coroutining is applied to the lexical
constraints.  As is well-known, memoizing parsers do not suffer from this
deficiency, and we present a memoizing interpreter below which does terminate.

\section{The Lemma Table proof procedure}

\newcommand{\C}{{\cal C}}

\noindent
This section presents a coroutining, memoizing CLP proof procedure.
The basic intuition behind our approach is quite natural in a CLP
setting like the one of H\"ohfeld and Smolka, which we sketch now.

A program is a set of definite clauses of the form
\[
p(X) \la q_1(X_1)\wedge \ldots\wedge q_n(X_n)\wedge \phi
\]
where the $X_i$ are vectors of variables, $p(X)$ and $q_i(X_i)$ are
relational atoms and $\phi$ is a basic constraint coming from a {\em
  basic constraint language} $\C$. $\phi$ will typically refer to some (or
all) of the variables mentioned. The language of basic constraints is
closed under conjunction and comes with (computable) notions of
consistency (of a constraint) and entailment
($\phi_1\models_{\C}\phi_2$) which have to be invariant under variable
renaming.%
\footnote{This essentially means that basic constraints can be recast as
  first-order predicates.}
Given a program $P$ and a goal $G$, which is a conjunction of relational atoms
and
constraints, a $P$-answer of $G$ is defined as {\em a consistent basic
  constraint $\phi$ such that $\phi\rightarrow G$ is valid in
every model of $P$}. SLD-resolution is generalized in this setting by
performing resolution only on relational atoms and simplifying
(conjunctions of) basic constraints thus collected in the goal list.
When finally only a consistent basic constraint remains, this is an
answer constraint $\phi$. Observe that this use of basic constraints
generalizes the use of substitutions in ordinary logic programming and
the (simplification of a) conjunction of constraints generalizes
unification. Actually, pure Prolog can be viewed as a syntactically
sugared variant of such a CLP language with equality constraints as basic
constraints, where a standard Prolog clause
%
\[
\rm p({\mit T}) \la q_1({\mit T_1}), \ldots, q_{\mit n}({\mit T_n})
\]
is seen as an abbreviation for a clause in which the equality constraints have
been made explicit by means of 
new variables and new equalities
\[
\begin{array}{rcl}
\rm p(X) & \la & X = {\mit T}, X_1 = {\mit T_1}, \ldots, X_{\mit n} = {\mit
T_n}, \\
      \rm & &       q_1(X_1), \ldots, q_{\mit n}(X_{\mit n}).
\end{array}
\]
Here the ${\rm X}_i$ are vectors of variables and the $T_i$ are vectors of
terms.

Now consider a standard memoizing proof procedure such as
Earley Deduction (Pereira and Warren~1983)\nocite{ParsingAsDeduction}
or the memoizing procedures described by Tamaki and Sato~(1986),
Vieille~(1989) or Warren~(1992) \nocite{v:2} \nocite{ts:oldt} \nocite{wa:mem}
 from this perspective.
Each memoized goal is associated with a set of bindings for its
arguments; so in CLP terms each memoized goal is a conjunction
of a single relational atom and zero or more equality constraints.
A completed (i.e., atomic) clause $\rm p({\mit T})$ with
an instantiated argument
$T$ abbreviates the non-atomic clause $\rm p(X) \la X = {\mit T}$,
where the equality constraint makes the instantiation specific.
Such equality constraints
are `inherited' via resolution by any clause that resolves with
the completed clause.

In the CLP perspective, variable-binding or equality constraints have
no special status; informally, all constraints can be treated in the
same way that pure Prolog treats equality constraints.  This is the
central insight behind the Lemma Table proof procedure: general
constraints are permitted to propagate into and out of subcomputations
in the same way that Earley Deduction
propagates variable bindings.
Thus the Lemma Table proof procedure generalizes
Earley Deduction in the following ways:
\begin{enumerate}
\item Memoized goals are in general conjunctions of relational atoms and
      constraints.  This allows constraints to be passed {\em into}
      a memoized subcomputation.

      We do not use this capability in the categorial grammar example (except
      to pass in variable bindings),
      but it is important in GB and HPSG parsing applications.
      For example, memoized goals in our GB parser consist of conjunctions
      of \xbar and ECP constraints.  Because the
      \xbar phrase-structure rules freely
      permit empty categories every string has infinitely many
      well-formed analyses that satisfy the \xbar constraints, but the
      conjoined ECP constraint rules out all but a very few of these empty
nodes.
%
\item Completed clauses can contain arbitrary negative literals
      (rather than just equality constraints, as in
      Earley Deduction).  This allows constraints to be passed
      {\em out of} a memoized subcomputation.

      In the categorial grammar example, the \verb+add_adjuncts+/2 and
      \verb+division+/2 associated with a lexical entry cannot be finitely
      resolved, as noted above, so e.g., a clause

\beginprolog
    x(\#X, [ontwijken], []) ::-
      [ add_adjuncts(s\verb+\+np\verb+\+np, X ) ].
\endprolog
      \noindent
      is classified as a completed clause; the \verb+add_adjuncts+/2 constraint
      in its body is inherited by any clause which uses this lemma.
\item Subgoals can be selected in any order (Earley Deduction always
      selects goals in left-to-right order).  This allows constraint
      coroutining {\em within} a memoized subcomputation.

      In the categorial grammar example, a category becomes more instantiated
      when it combines with arguments, allowing eventually the
      \verb+add_adjuncts+/2 and
      \verb+division+/2 to be deterministically resolved.  Thus we use the
      flexibility in the selection of goals to run constraints whenever their
      arguments are sufficiently instantiated, and delay them otherwise.
\item Memoization can be selectively applied (Earley Deduction memoizes
      every computational step).  This can significantly improve
      overall efficiency.

      In the categorial grammar example only \verb+x+/3 goals are memoized
      (and thus only these goals incur the cost of table management).
\end{enumerate}

\noindent
The `abstraction' step, which is
used in most memoizing systems (including complex feature grammar
chart parsers where it is somewhat confusingly called `restriction',
as in Shieber~1985)\nocite{Shieber85}, receives an elegant treatment in a CLP
approach;
an `abstracted' goal is merely one in which not all of the equality
constraints associated with the variables appearing in the goal are
selected with that goal.%
\footnote{After this paper was accepted, we discovered that a more general
formulation of abstraction is required for systems using a hierarchy of
types,
such as typed feature structure constraints~(Carpenter~1992).
In applications of the Lemma Table Proof Procedure to such systems
it may be desirable to abstract from a `strong' type constraint in the
body of a clause to a logically `weaker' type constraint in the memoized
goal.  Such a form of abstraction cannot be implemented using the
selection rule alone.}

For example, because of the backward application rule
and the left-to-right evaluation our parser uses, eventually it will
search at every left string position for an uninstantiated category
(the variable \verb+Y+ in the clause), we might as well abstract
all memoized goals of the form $\verb+x(+\it C, L, R\verb+)+$ to
$\verb+x(_+\it , L, \verb+_)+$, i.e., goals in which the category and
right string position are uninstantiated.  Making the equality constraints
explicit, we see that the abstracted goal is obtained by merely selecting
the underlined subset of these below:
\[
\rm \underline{\verb+x(+ X_1, X_2, X_3\verb+)+},  X_1 = {\mit C},
        \underline{X_2 = {\mit L}}, X_3 = {\mit R}.
\]
While our formal presentation does not discuss
abstraction (since it can be implemented in terms of constraint selection
as just described), because our implementation
uses the underlying Prolog's unification mechanism to solve
equality constraints over terms, it provides an explicit abstraction operation.

Now we turn to the specification of the algorithm itself,
beginning with the basic computational entities it uses.

\begin{Definition} \rm
A (generalized) {\em goal} is a multiset of relational atoms and constraints.
A (generalized) {\em clause} $\clause{H_0}{B_0}$ is an ordered
pair of generalized goals, where
$H_0$ contains at least one relational atom.  A relational
interpretation $\A$ (see H\"ohfeld and Smolka~1988 for definition)
satisfies a goal $G$ iff $\A$ satisfies each element of $G$, and
it satisfies a clause
$\clause{H_0}{B_0}$ iff either $\A$ fails to satisfy
some element of $B_0$ or $\A$ satisfies each element of $H_0$.
\end{Definition}

\noindent
This generalizes the standard notion of clause by allowing the
head $H_0$ to consist of more than one atom.  The head $H_0$ is
interpreted conjunctively; i.e., if each element of $B_0$ is
true, then so is each element of $H_0$.  The standard definition
of resolution extends unproblematically to such clauses.

\begin{Definition} \rm
We say that a clause {\em $\cvar{c}_0  =\clause{H_0}{B_0}$
resolves with a clause $\cvar{c}_1  =\clause{H_1}{B_1}$
on a non-empty set of literals $C\subseteq B_0$} 
iff
there is a variant ${\cvar{c}_1}'$ of $\cvar{c}_1$ of
the form $\clause{C}{{B_1}'}$ such that
$V(\cvar{c}_0) \cap V({B_1}') \subseteq V(C)$ (i.e.,
the variables common to $\cvar{c}_0$ and ${B_1}'$ also
appear in C, so there is no accidental variable sharing).

If $\cvar{c}_0$ resolves with $\cvar{c}_1$ on $C$, then
the clause $\clause{H_0}{(B_0 - C) \cup {B_1}'}$ is
called a {\em resolvent of $\cvar{c}_0$ with $\cvar{c}_1$ on $C$}.
\end{Definition}

\noindent
Now we define items, which are
the basic computational units that appear on the
agenda and in the lemma tables, which record
memoized subcomputations.

\begin{Definition} \rm
An {\em item} is a pair $\langle t, \cvar{c} \rangle$ where
$\cvar{c}$ is a clause and
$t$ is a tag, i.e., one of $\program$, $\solution$ or
$\tabelle(B)$ for some goal $B$.
A {\em lemma table for a goal $G$} is a pair $\langle G, L_G \rangle$
where $L_G$ is a finite list of items.
\end{Definition}

\noindent
The algorithm manipulates a set $T$ of lemma tables which has
the property that the first components of any two distinct members
of $T$ are distinct.
This justifies speaking of the (unique) lemma table in $T$ for
a goal $G$.

Tags are associated with clauses by a user-specified control rule,
as described below.
The tag associated with a clause in an item identifies the operation
that should be performed on that clause.  The $\solution$ tag labels
`completed' clauses, 
the $\program$ tag directs the proof procedure to perform a non-memoizing
resolution of one of the clause's negative literals with program clauses
(the particular negative literal is chosen by a user-specified selection rule,
as in standard SLD resolution),
and the $\tabelle(B)$ tag indicates that a subcomputation with root goal $B$
(which is always a subset of the clause's negative
literals) should be started.

\begin{Definition} \rm
A {\em control rule} is a function from clauses $\clause{G}{B}$ to
one of $\program$, $\solution$ or $\tabelle(C)$ for some goal $C \subseteq B$.
A {\em selection rule} is a function from clauses $\clause{G}{B}$
where $B$ contains at least one relational atom to relational atoms $a$,
where $a$ appears in $B$.
\end{Definition}

\noindent
Because $\program$ steps do not require memoization
and given the constraints on the control rule just mentioned,
the list $L_G$ associated with a
lemma table  $\langle G, L_G \rangle$ will only contain
items of the form $\langle t, \clause{G}{B} \rangle$
where $t$ is either $\solution$ or
$\tabelle(C)$ for some goal $C \subseteq B$.

\begin{Definition} \rm
To {\em add an item an item $e = \langle t, \clause{H}{B} \rangle$ to
its table} means to replace the table $\langle H, L \rangle$ in $T$
with $\langle H, [e | L] \rangle$.
\end{Definition}

\newlength{\figwidth}
\setlength{\figwidth}{\textwidth}
\addtolength{\figwidth}{-10pt}
\begin{figure*}[hft]
\begin{center}
\fbox{
\begin{minipage}{\figwidth}
\begin{description}
\item[Input] A non-empty goal $G$, a program $P$, a selection rule $S$,
and a control rule $R$.
\item[Output] A set of goals $G'$ for which
  $R(G') = \solution$ and
  $ P \, \models \, G \leftarrow G'$.
\item[Global Data Structures] A set $T$ of lemma tables and a set $A$
  of items called the agenda.
\item[Algorithm]
  Set $T := \{ \langle G, \emptyset\rangle \} $ and
  $A := \{ \langle program, \clause{G}{G} \rangle \} $.

  Until $A$ is empty, do:

  Remove an item $e = \langle t, \cvar{c} \rangle$ from $A$.\hfil\break
  Case $t$ of
  \begin{description}
  \item[\program]
    For each clause $\cvar{p} \in P$ such that
    $\cvar{c}$ resolves with $\cvar{p}$ on
    $S(\cvar{c})$, choose a corresponding resolvent
    $\cvar{c'}$ and add $\langle R(\cvar{c'}),\cvar{c'} \rangle$ to
    $A$.
  \item[\tabelle($B$)]
    Add $e$ to its table.$^3$

    If $T$ contains a table $\langle B', L \rangle$ where $B'$ is a variant
    of $B$ then for each item $\langle \solution, \cvar{c'}
    \rangle \in L$ such that $\cvar{c}$ resolves with $c'$ on $B$
    choose a corresponding resolvent $c''$ and add $\langle R(c''),
    c'' \rangle $ to $A$.

    Otherwise, add a new table $\langle B, \emptyset \rangle$ to $T$,
    and add $\langle \program, \clause{B}{B} \rangle$ to the agenda.
  \item[\solution]
    Add $e$ to its table.

    Let $\cvar{c} = \clause{H}{B}$. Then for each item of the form
    $\langle \tabelle{(H')}, c' \rangle$ in any table in $T$ where
    $H'$ is a variant of $H$ and $c'$ resolves with $c$ on $H'$,
    choose a corresponding resolvent $c''$ and add $\langle R(c''),
    c'' \rangle $ to $A$.
  \end{description}

  Set $ \Gamma := \{ B : \langle \solution, \clause{G}{B} \rangle \in
  L, \, \langle G, L \rangle \in T \} $.

\end{description}
\end{minipage}
}
\vspace*{-\abovedisplayskip}
\end{center}
\caption{The Lemma Table algorithm}
\label{fig:ltpp}
\end{figure*}

\noindent
The formal description of the Lemma Table proof procedure is given in
Figure~\ref{fig:ltpp}.
We prove the soundness and completeness of the proof procedure in
D\"orre and Johnson~(in preparation).
In fact, soundness is easy to show, since
all of the operations are resolution steps. 
Completeness follows
 from the fact that Lemma Table proofs can be `unfolded' into standard
SLD search trees (this unfolding is well-founded because the first
step of every $\tabelle$-initiated subcomputation is required to be a
$\program$ resolution), so completeness follows from H\"ohfeld and
Smolka's completeness theorem for SLD resolution in CLP.
\footnotetext[3]{
In order to handle the more general form of abstraction discussed in
footnote 2 which may be
useful with typed feature structure constraints,
replace $B$ with $\alpha(B)$ in this step,
where $\alpha(B)$ is the result of applying the abstraction operation to $B$.

The abstraction operation should have the property that $\alpha(B)$ is exactly
the same as $B$, except that zero or more constraints
in $B$ are replaced with logically weaker constraints.}

\section{A worked example}
\noindent
Returning to the categorial grammar example above, the control rule and
selection
rule are specified by the Prolog code below, which can be
informally described as follows.   All \verb+x+/3 literals are classified as
`memo' literals, and \verb+add_adjuncts+/2 and \verb+division+/2 whose second
arguments are not sufficiently instantiated are classified as `delay' literals.
If the clause contains a memo literal $G$, then the control rule returns
$\tabelle{([G])}$.
Otherwise, if the clause contains any non-delay literals, then the control
rule returns $\program$ and the selection rule chooses the left-most
such literal.
If none of the above apply, the control rule returns $\solution$.  To simplify
the interpreter code, the
Prolog code for the selection rule and $\tabelle{(G)}$ output of the control
rule
also return the remaining literals along with chosen goal.
\beginprolog
:- ensure_loaded(library(lists)).
:- op(990, fx, [delay, memo]).

delay division(_, X/Y) :- var(X), var(Y).
delay add_adjuncts(_, X/Y) :- var(X), var(Y).

memo x(_,_,_).

control(Gs0, Control) :-
    memo(G), select(G, Gs0, Gs)
        -> Control = table([G], Gs) ;
    member(G, Gs0), \verb+\++ delay(G)
        -> Control = program ;
    Control = solution.

selection(Gs0, G, Gs) :-
    select(G1, Gs0, Gs1), \verb-\-+ delay(G1)
        -> G = G1, Gs = Gs1.
\endprolog
\noindent
Because we do not represent
variable binding as explicit constraints, we cannot
implement `abstraction' by means of the control rule and require
an explicit abstraction operation.  The abstraction operation here
unbinds the first and third arguments of \verb+x+/3 goals, as
discussed above.
\beginprolog
abstraction([x(_,Left,_)], [x(_,Left,_)]).
\endprolog
\begin{figure*}
{\catcode`_=\active
\def\Box{\raise.4ex\hbox{$\scriptstyle\#$}}
\ltprogram{0}{1}{0}{\ltbclause{\ltatom{x(A,[l_t,o],B)}}{\ltulatom{x(A,[l_t,o],B)}}}
\lttable{0}{2}{1}{\ltbclause{\ltatom{x(A,[l_t,o],B)}}{\ltulatom{x(A/C,[l_t,o],D)}, \ltatom{x(C,D,B)}}}
\lttable{0}{3}{1}{\ltbclause{\ltatom{x(A,[l_t,o],B)}}{\ltulatom{x(C,[l_t,o],D)}, \ltatom{x(A \backslash  C,D,B)}}}
\ltprogram{0}{4}{1}{\ltbclause{\ltatom{x(A,[l_t,o],[o])}}{\ltulatom{lex(l_t,A)}}}
\ltsolution{0}{5}{4}{\ltbclause{\ltatom{x(A/\Box
B,[l_t,o],[o])}}{\ltatom{add(s \backslash  np/(s \backslash  np),C)},
\ltatom{div(C,A/B)}}}
\lttable{0}{6}{2,5}{\ltbclause{\ltatom{x(A,[l_t,o],B)}}{\ltatom{add(s
\backslash  np/(s \backslash  np),C)}, \ltatom{div(C,A/D)}, \ltulatom{x(\Box
D,[o],B)}}}
\ltprogram{1}{7}{6}{\ltbclause{\ltatom{x(A,[o],B)}}{\ltulatom{x(A,[o],B)}}}
\lttable{1}{8}{7}{\ltbclause{\ltatom{x(A,[o],B)}}{\ltulatom{x(A/C,[o],D)},
\ltatom{x(C,D,B)}}}
\lttable{1}{9}{7}{\ltbclause{\ltatom{x(A,[o],B)}}{\ltulatom{x(C,[o],D)},
\ltatom{x(A \backslash  C,D,B)}}}
\ltprogram{1}{10}{7}{\ltbclause{\ltatom{x(A,[o],[])}}{\ltulatom{lex(o,A)}}}
\ltsolution{1}{11}{10}{\ltbclause{\ltatom{x(\Box  A,[o],[])}}{\ltatom{add(s
\backslash  np \backslash  np,A)}}}
\ltsolution{0}{12}{6,11}{\ltbclause{\ltatom{x(A,[l_t,o],[])}}{\ltatom{add(s
\backslash  np \backslash  np,B)}, \ltatom{add(s \backslash  np/(s \backslash
np),C)}, \ltatom{div(C,A/B)}}}
\lttable{0}{13}{2,12}{\ltbclause{\ltatom{x(A,[l_t,o],B)}}{\ltatom{add(s
\backslash  np \backslash  np,C)}, \ltatom{add(s \backslash  np/(s \backslash
np),D)}, \ltatom{div(D,A/E/C)}, \ltulatom{x(E,[],B)}}}
\ltprogram{2}{14}{13}{\ltbclause{\ltatom{x(A,[],B)}}{\ltulatom{x(A,[],B)}}}
\lttable{2}{15}{14}{\ltbclause{\ltatom{x(A,[],B)}}{\ltulatom{x(A/C,[],D)},
\ltatom{x(C,D,B)}}}
\lttable{2}{16}{14}{\ltbclause{\ltatom{x(A,[],B)}}{\ltulatom{x(C,[],D)},
\ltatom{x(A \backslash  C,D,B)}}}
\lttable{0}{17}{3,12}{\ltbclause{\ltatom{x(A,[l_t,o],B)}}{\ltatom{add(s
\backslash  np \backslash  np,C)}, \ltatom{add(s \backslash  np/(s \backslash
np),D)}, \ltatom{div(D,E/C)}, \ltulatom{x(A \backslash  E,[],B)}}}
\lttable{1}{18}{9,11}{\ltbclause{\ltatom{x(A,[o],B)}}{\ltatom{add(s \backslash
np \backslash  np,C)}, \ltulatom{x(A \backslash  \Box  C,[],B)}}}
\lttable{0}{19}{3,5}{\ltbclause{\ltatom{x(A,[l_t,o],B)}}{\ltatom{add(s
\backslash  np/(s \backslash  np),C)}, \ltatom{div(C,D/E)}, \ltulatom{x(A
\backslash (D/\Box  E),[o],B)}}}
}
\caption{The items produced during the proof of
{\tt x(C,[lijkt{\ul}te,ontwijken],{\ul})} using the control and selection rules
specified in the text.
The prefix $t.n[a]\;T$ identifies the table $t$ to which this item belongs,
assigns this item a unique identifying number $n$, provides the number(s) of
the
item(s) $a$ which caused this item to be created, and displays its tag $T$
({\protect\footnotesize P}
for `program', {\protect\footnotesize T} for `table' and
{\protect\footnotesize S} for `solution').
The selected literal(s) are shown underlined.
To save space, `add{\ul}adjuncts' is abbreviated by `add', `division' by `div',
`lijkt{\ul}te' by `lt', and `ontwijken' by `o'. \label{fig:trace}}
\end{figure*}
\noindent
Figure~\ref{fig:trace} depicts the proof
of a parse of the verb cluster in (1).
Item~1 is generated by the initial goal; its sole negative literal is selected
for program resolution, producing items~2--4 corresponding to three program
clauses
for \verb+x+/3.  Because items 2 and 3 contain `memo' literals, the
control rule tags them $\tabelle$; there already is a table for
a variant of these goals (after abstraction).
Item~4 is tagged $\program$ 
because it contains a
negative literal that is not `memo' or `delay'; the resolution of this
literal with the program clauses for \verb+lex+/3 produces item~5 containing
the constraint literals associated with {\em lijkt te}.
Both of these are classified as `delay'
literals, so item~5 is tagged $\solution$, and both are `inherited'
when item~5 resolves with the $\tabelle$-tagged items~2 and~3,
producing items~6 (corresponding to a right application analysis with
{\em lijkt te} as functor) and item~19 (corresponding to a left application
analysis
with {\em ontwijken} as functor) respectively.  Item~6 is tagged $\tabelle$,
since
it contains a \verb+x+/3 literal; because this goal's second argument (i.e.,
the
left string position) differs from that of the goal associated with table~0, a
new table (table~1) is constructed, with item~7 as its first item.

The three program clauses for  \verb+x+/3 are used to resolve the selected
literal
in item~7, just as in item~1, yielding items~8--10.  The \verb+lex+/3 literal
in item~10 is resolved with the appropriate program clause, producing item~11.
Just as in item~5, the second argument of the single literal in item~11 is
not sufficiently instantiated, so item~11 is tagged $\solution$,
and the unresolved literal is `inherited' by item~12.  Item~12 contains the
partially resolved analysis of the verb complex.
Items~13--16 analyze the empty string;
notice that there are no $\solution$ items for table~2.  Items~17--19
represent partial alternative analyses of the verb cluster where the
two verbs combine using other rules than forward application; again,
these yield no $\solution$ items, so item~12 is the sole analysis of
the verb cluster.

\section{A simple interpreter}
\noindent
This section describes an implementation of the Lemma Table proof procedure
in Prolog, designed for simplicity rather than efficiency.  Tables are stored
in the Prolog database, and no explicit agenda is used.
The dynamic predicate $\verb+goal_table(+G, I\verb+)+$
records the initial goals $G$ for each table subcomputation and that table's
identifying index $I$ (a number assigned to each table when it is created).
The dynamic predicate $\verb+table_solution(+I, S\verb+)+$ records all of
the $\solution$ items generated for table~$I$ so far, and
$\verb+table_parent(+I, T\verb+)+$ records the $\tabelle$ items $T$,
called `parent items' below, which
are `waiting' for additional $\solution$ items from table~$I$.

The `top level' goal is $\verb+prove(+G,Cs\verb+)+$, where $G$ is a single
atom (the goal to be proven), and $Cs$ is a list of (unresolved) solution
constraints
(different solutions are enumerated through backtracking).  \verb+prove+/2
starts by retracting the tables associated with previous computations,
asserting the table entry associated with the initial goal, and then
calls \verb+take_action+/2 to perform a program resolution on the initial
goal.  After all succeeding steps are complete, \verb+prove+/2 returns
the solutions associated with table~0.
\beginprolog
prove(Goal, _Constraints) :-
    retractall(goal_table(_,_)),
    retractall(table_solution(_,_)),
    retractall(table_parent(_, _)),
    retractall(counter(_)),
    assert(goal_table([Goal], 0)),
    take_action(program, [Goal]::-[Goal], 0),
    fail.
prove(Goal, Constraints) :-
    table_solution(0, [Goal]::-Constraints).
\endprolog
\noindent
The predicate $\verb+take_action(+L, C, I\verb+)+$ processes items.
$L$ is the item's label, $C$ its clause and $I$ is the index of the
table it belongs to.  The first clause calls \verb+complete+/2 to
resolve the $\solution$ clause with any parent items the table may
have, and the third clause constructs a parent item term (which
encodes both the clause, the tabled goal, and the index of the
table the item belongs to) and calls \verb+insert_into_table+/2
to insert it into the appropriate table.
\beginprolog
take_action(solution, Clause, Index) :-
    assert(table_solution(Index, Clause)),
    findall(P, table_parent(Index, P),
            ParentItems),
    member(ParentItem, ParentItems),
    complete(ParentItem, Clause).
take_action(program, Head::-Goal, Index) :-
    selection(Goal, Selected, Body1),
    Selected ::- Body0,
    append(Body0, Body1, Body),
    control(Body, Action),
    take_action(Action, Head::-Body, Index).
take_action(table(Goal,Other), Head::-_Body,
            Index) :-
    insert_into_table(Goal,
        tableItem(Head, Goal, Other, Index)).
\endprolog
\noindent
\verb+complete+/2 takes an item labeled $\tabelle$ and a clause,
resolves the head of the clause with the item, and calls
\verb+control+/2 and \verb+take_action+/3 to process the resulting
item.
\beginprolog
complete(tableItem(Head, Goal, Body1, Index),
         Goal::-Body0) :-
    append(Body0, Body1, Body),
    control(Body, Action),
    take_action(Action, Head::-Body, Index).
\endprolog
\noindent
The first clause \verb+insert_into_table+/2 checks to see if a
table for the goal to be tabled has already been constructed
(\verb+numbervars+/3 is used to ground a copy of the term).
If an appropriate table does not exist, the second clause
calls \verb+create_table+/3 to construct one.
\beginprolog
insert_into_table(Goal, ParentItem) :-
    copy_term(Goal, GoalCopy),
    numbervars(GoalCopy, 0, _),
    goal_table(GoalCopy, Index),
    !,
    assert(table_parent(Index, ParentItem)),
    findall(Sol, table_solution(Index, Sol),
                 Solutions), !,
    member(Solution, Solutions),
    complete(ParentItem, Solution).
insert_into_table(Goal0, ParentItem) :-
    abstraction(Goal0, Goal), !,
    create_table(Goal, ParentItem, Index),
    take_action(program, Goal::-Goal, Index).
\endprolog
\noindent
\verb+create_table+/3 performs the necessary database manipulations
to construct a new table for the goal, assigning a new index for
the table, and adding appropriate entries to the indices.

\medskip
\beginprolog
create_table(Goal, ParentItem, Index) :-
    (retract(counter(Index0)) -> true
    ; Index0=0),
    Index is Index0+1,
    assert(counter(Index)),
    assert(goal_table(Goal, Index)),
    assert(table_parent(Index, ParentItem)).
\endprolog

\section{Conclusion}
\noindent
This paper has presented a general framework which allows both
constraint coroutining and memoization.  To achieve maximum generality
we stated the Lemma Table proof procedure in H\"ohfeld and
Smolka's~(1988) CLP framework, but the basic idea---that arbitrary
constraints can be allowed to propagate in essentially the same way
that variable bindings do---can be applied in most approaches to
complex feature based parsing.  For example, the technique can be
used in chart parsing: in such a system an edge consists not only
of a dotted rule and associated variable bindings (i.e., instantiated
feature terms), but also contains
zero or more as yet unresolved constraints that are propagated (and
simplified if sufficiently instantiated) during application of
the fundamental rule.

At a more abstract level, the identical propagation of both
variable bindings and more general constraints leads us to question whether
there is any principled difference between them.
While still preliminary, our research suggests that it is often
possible to reexpress complex feature based grammars more succinctly
by using more general constraints.



\begin{thebibliography}{AI}

\bibitem[BoumaNoord~94]{BoumaNoord94a}
G.~Bouma and G.~van Noord.
\newblock Constraint-Based Categorial Grammar.
\newblock In {\em Proceedings of the 32nd Annual Meeting of the ACL, New Mexico
  State University}, Las Cruces, New Mexico, 1994.

\bibitem[Carpenter~92]{Carpenter}
B.~Carpenter.
\newblock The Logic of Typed Feature Structures.
\newblock Cambridge Tracts in Theoretical Computer Science~32.
Cambridge University Press. 1992.

\bibitem[D\"orre~93]{DorreDyana}
J.~D\"orre.
\newblock Generalizing {E}arley deduction for constraint-based grammars.
\newblock In J.~D{\"o}rre (ed.), {\em Computational Aspects of
  Constraint-Based Linguistic Description I, DYANA-2 deliverable R1.2.A}.
  ESPRIT, Basic Research Project 6852, July 1993.

\bibitem[D\"orre~inprep]{DorreInprep}
J.~D\"orre and M. Johnson.
\newblock Memoization and coroutined constraints.
\newblock ms. Institut f\"ur maschinelle Sprachverarbeitung,
Universit\"at Stuttgart.

\bibitem[H\"ohfeldSmolka~88]{DR88}
M.~H\"ohfeld and G.~Smolka.
\newblock Definite Relations over Constraint Languages.
\newblock LILOG Report~53, IWBS, IBM Deutschland, Postfach 80 08 80, 7000
  Stuttgart 80, W. Germany, October 1988. (available on-line by
  anonymous ftp from /duck.dfki.uni--sb.de:/pub/papers)

\bibitem[Johnson~93]{MJ93}
M.~Johnson.
\newblock Memoization in Constraint Logic Programming.
\newblock Presented at {\em First Workshop on Principles and Practice of
Constraint Programming, April 28--30 1993,
Newport, Rhode Island.}

\bibitem[PereiraWarren~83]{ParsingAsDeduction}
F.~C. Pereira and D.~H. Warren.
\newblock Parsing as Deduction.
\newblock In {\em Proceedings of the 21st Annual Meeting of the ACL,
  Massachusetts Institute of Technology}, pp. 137--144, Cambridge, Mass., 1983.

\bibitem[Shieber~85]{Shieber85}
S.~M. Shieber.
\newblock Using Restriction to Extend Parsing Algorithms for
  Complex-Feature-Based Formalisms.
\newblock In {\em Proceedings of the 23rd Annual Meeting of the Association for
  Computational Linguistics}, pp. 145--152, 1985.

\bibitem[Tamaki~1986]{ts:oldt} Tamaki, H. and T. Sato. ``OLDT resolution with
tabulation'', in {\em Proceedings of Third International Conference
on Logic Programming}, Springer-Verlag, Berlin, pages~84--98.  1986.

\bibitem[Vieille~1989]{v:2} Vieille, L.
``Recursive query processing: the power of logic'',
Theoretical Computer Science 69, pages~1--53.  1989.

\bibitem[Warren~1992]{wa:mem} Warren, D. S.
``Memoing for logic programs'', in
{\em Communications of the ACM} 35:3, pages~94--111. 1992.

\end{thebibliography}

\end{document}